%% file: main.tex
\documentclass[letterpaper, 10 pt, conference]{ieeeconf}  

\IEEEoverridecommandlockouts
\title{\LARGE 
Semantic Segmentation of Underwater Imagery: Dataset and Benchmark\textbf{*}
}

\author{
    Md Jahidul Islam$^1$, Chelsey Edge$^2$, Yuyang Xiao$^3$, Peigen Luo$^4$, Muntaqim Mehtaz$^5$, \\ Christopher Morse$^6$, Sadman Sakib Enan$^7$ and Junaed Sattar$^8$% <-this % stops a space
    \thanks{The authors are with the Interactive Robotics and Vision Laboratory (IRVLab), Department of Computer Science and Engineering (CSE), Minnesota Robotics Institute (MnRI), University of Minnesota, Twin Cities, US. {\tt\footnotesize \{$^1$islam034, $^2$edge0037, $^3$xiao0153, $^4$luo00034, $^5$mehta216, $^6$morse164, $^7$enan0001, $^8$junaed\}@umn.edu}
    }
    \thanks{\textbf{* }This pre-print is accepted for publication at the IROS 2020; find more information at \url{https://github.com/xahidbuffon/SUIM-Net}.}
}
\usepackage{epsfig}
\usepackage{graphicx}
\usepackage{amsmath}
\usepackage{amssymb}
\usepackage{paralist}
\usepackage{multicol}
\usepackage{multirow}
\usepackage{color}
\usepackage{subcaption}
\newcommand{\ie}{\textit{i.e.}}
\newcommand{\eg}{\textit{e.g.}}
 
\newcommand*\rot{\rotatebox{90}}

\usepackage[table]{xcolor}% http://ctan.org/pkg/xcolor
\usepackage{url}
\usepackage[colorlinks,bookmarksopen,bookmarksnumbered,citecolor=blue,urlcolor=blue]{hyperref}
\definecolor{black}{rgb}{0,0,0}
\definecolor{blue}{rgb}{0,0,1}
\definecolor{green}{rgb}{0,1,0}
\definecolor{sky}{rgb}{0,1,1}
\definecolor{red}{rgb}{1,0,0.}
\definecolor{pink}{rgb}{1,0,1}
\definecolor{yellow}{rgb}{1,1,0}
\definecolor{white}{rgb}{1,1,1}
\begin{document}

\maketitle
\thispagestyle{empty}
\pagestyle{empty}

\input{src/abstract.tex}

\input{src/intro}

\input{src/suim}

\input{src/rel}
\input{src/suimnet}

\input{src/benchmark}
\input{src/con}

% TODO: adjust this before submission
\addtolength{\textheight}{-0cm} 

\bibliographystyle{IEEEtran}
\bibliography{refs.bib}

\end{document}

%% file: src/abstract.tex
\begin{abstract}
In this paper, we present the first large-scale dataset for semantic Segmentation of Underwater IMagery (SUIM). It contains over 1500 images with pixel annotations for eight object categories: fish (vertebrates), reefs (invertebrates), aquatic plants, wrecks/ruins, human divers, robots, and sea-floor. The images have been rigorously collected during oceanic explorations and human-robot collaborative experiments, and annotated by human participants. We also present a benchmark evaluation of state-of-the-art semantic segmentation approaches based on standard performance metrics. In addition, we present SUIM-Net, a fully-convolutional encoder-decoder model that balances the trade-off between performance and computational efficiency. It offers competitive performance while ensuring fast end-to-end inference, which is essential for its use in the autonomy pipeline of visually-guided underwater robots. In particular, we demonstrate its usability benefits for visual servoing, saliency prediction, and detailed scene understanding. With a variety of use cases, the proposed model and benchmark dataset open up promising opportunities for future research in underwater robot vision.
\end{abstract}

%% file: src/intro.tex
\section{Introduction}
Semantic segmentation is a well-studied problem in the domains of robot vision and deep learning~\cite{garcia2017review,chen2017deeplab,badrinarayanan2015segnet} for its usefulness in estimating scene geometry, inferring interactions and spatial relationships among objects, salient object identification, and more. 
It is particularly important for detailed scene understanding in autonomous driving by visually-guided robots. Over the last decade, substantial contributions from both industrial and academic researchers have led to remarkable advancements of the state-of-the-art (SOTA) methodologies for semantic segmentation~\cite{garcia2017review,long2015fully}. This success is largely propelled by various genres of deep convolutional neural network (CNN)-based models that learn from large collections of annotated data~\cite{ronneberger2015u,chen2017rethinking}. Several large-scale benchmark datasets of terrestrial imagery~\cite{oquab2014learning,lin2014microsoft} and videos~\cite{perazzi2016benchmark} provide a standard platform for such research and fuel the rapid development of the relevant literature. To date, the SOTA semantic segmentation models are core elements in the visual perception pipelines of most terrestrial robots and systems.

\begin{figure}[t]
    \centering
    \begin{subfigure}{0.5\textwidth} 
    \centering
        \includegraphics[width=0.98\linewidth]{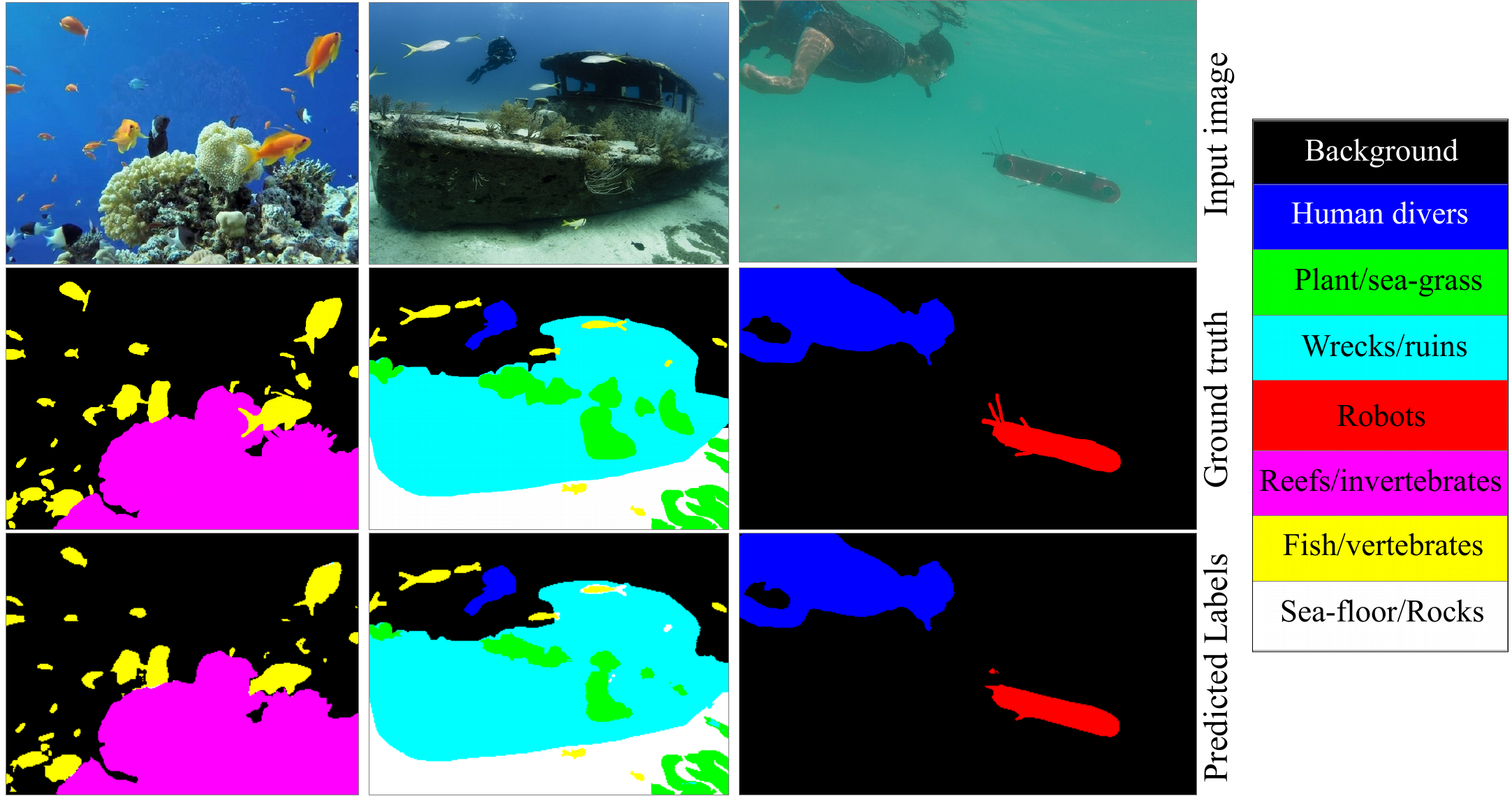}
        \caption{Three instances of semantic segmentation by SUIM-Net and respective ground truth labels are shown; the object categories and color codes are provided on the right.}
        \label{fig_1a}
    \end{subfigure}
    \vspace{2mm}
    
    \begin{subfigure}{0.5\textwidth} 
    \centering
        \includegraphics[width=0.98\linewidth]{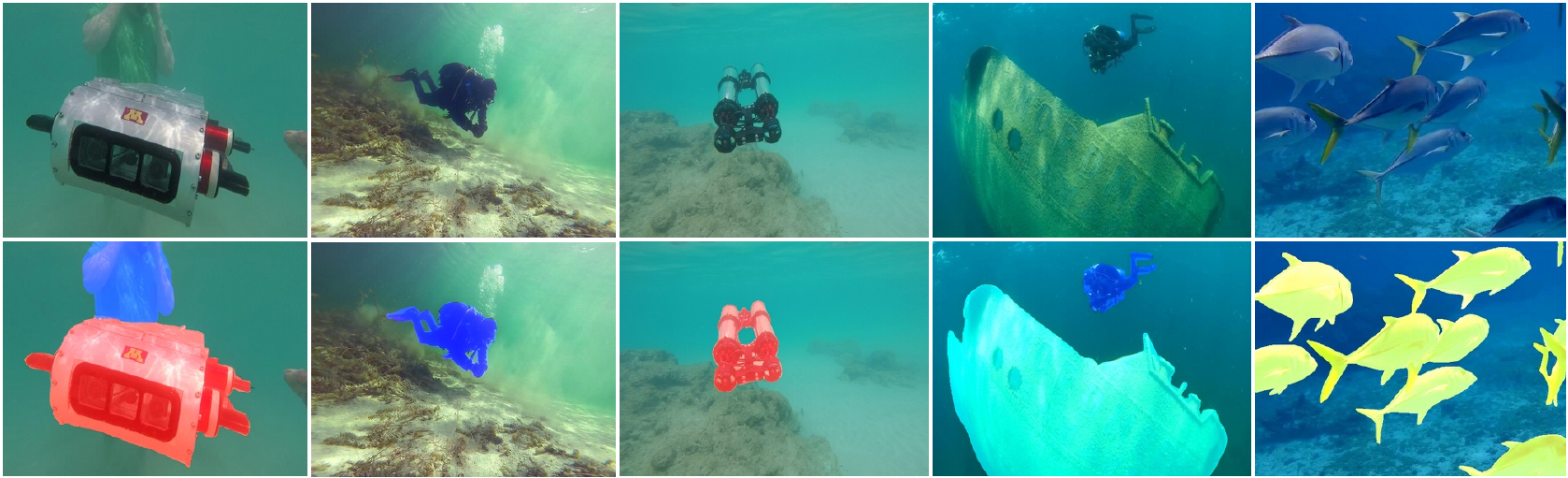}
        \caption{Pixel-level detection for specific object categories: humans, robots, fish, and wrecks/ruins are shown (the segmentation masks are overlayed on the bottom row images); such fine-grained object localization is useful for visual attention modeling and servoing.}
        \label{fig_1b}
    \end{subfigure}
    \caption{Demonstrations for semantic segmentation of underwater scenes and other use cases of the proposed SUIM dataset and SUIM-Net model.}
    \label{fig:intro}
\end{figure}

For visually-guided underwater robots, however, the existing solutions for semantic segmentation and scene parsing are significantly less advanced. The practicalities and limitations are twofold. 
First, the visual content of underwater imagery is entirely different because of the domain-specific object categories, background patterns, and optical distortion artifacts~\cite{islam2019fast}; hence, the learning-based SOTA models trained on terrestrial data are not directly applicable. 
Secondly, there are no underwater datasets to facilitate large-scale training and benchmark evaluation of semantic segmentation models for general-purpose use. The existing large-scale annotated data and relevant methodologies are tied to specific applications such as coral-reef classification and coverage estimation~\cite{beijbom2012automated,alonso2019coralseg,VAIME}, fish detection and segmentation~\cite{ravanbakhsh2015automated,chuang2011automatic}, etc. 
Other datasets contain either binary annotations for salient foreground pixels~\cite{islam2020sesr} or semantic labels for very few object categories (\eg, sea-grass, rocks/sand, etc.)~\cite{zhou2019underwater}. Therefore, the large-scale learning-based semantic segmentation methodologies for underwater imagery are not explored in depth in the literature.     
Besides, the traditional class-agnostic approaches are only suitable for simple tasks such as foreground segmentation~\cite{li2016mapreduce,padmavathi2010non}, obstacle detection~\cite{arain2019improving}, saliency prediction~\cite{zhu2017underwater}, etc.; they are not generalizable for multi-object semantic segmentation.

We attempt to address these limitations by presenting a large-scale annotated dataset for semantic segmentation of underwater scenes in general-purpose robotic applications. 
As shown in Figure~\ref{fig:intro}, the proposed SUIM dataset considers object categories for fish, reefs, aquatic plants, and wrecks/ruins, which are of primary interest in many underwater exploration and surveying applications~\cite{girdhar2014autonomous,shkurti2012multi,bingham2010robotic}. Additionally, it contains pixel annotations for human divers, robots/instruments, and sea-floor/rocks; these are major objects of interest in human-robot cooperative applications~\cite{islam2018understanding,sattar2009vision}. The SUIM dataset contains $1525$ natural underwater images and their ground truth semantic labels; it also includes a test set of $110$ images. The dataset and relevant resources are available at \url{https://irvlab.cs.umn.edu/resources/suim-dataset}.

Moreover, we conduct a thorough experimental evaluation of several SOTA semantic segmentation models and compare their performance on the SUIM dataset. We also design a novel encoder-decoder model named SUIM-Net, which offers much faster run-time than the SOTA models while achieving competitive semantic segmentation performance. In addition to presenting the conceptual model and detailed network architecture of SUIM-Net, we analyze its performance in quantitative and qualitative terms based on standard metrics. Furthermore, we demonstrate various use cases of SUIM-Net and specify corresponding training configurations for the SUIM dataset. 
The model and associated training pipelines are released for academic research at \url{http://irvlab.cs.umn.edu/image-segmentation/suim-and-suim-net}.

%% file: src/suim.tex
\section{The SUIM Dataset}
We consider the following object categories for semantic labeling in the SUIM dataset: \textit{a)} waterbody background ({\tt BW}), \textit{b)} human divers ({\tt HD}), \textit{c)} aquatic plants/flora ({\tt PF}), \textit{d)} wrecks/ruins ({\tt WR}), \textit{e)} robots and instruments ({\tt RO}), \textit{f)} reefs and other invertebrates ({\tt RI}), \textit{g)} fish and other vertebrates ({\tt FV}), and \textit{h)} sea-floor and rocks ({\tt SR}). 
As depicted in Table~\ref{tab:my_label}, we use 3-bit binary RGB colors to represent these eight object categories in the image space.

\begin{table}[ht]
    \centering
    \caption{The object categories and corresponding color codes for pixel annotations in the SUIM dataset.}
    \begin{tabular}{l||c|r}
      \hline
      \textbf{Object category} & {\textbf{RGB color}} & {\textbf{Code}} \\ \hline \hline
      Background (waterbody) & \cellcolor{black}{\tt {\color{white}000}} & {\tt BW} \\ \hline
      Human divers & \cellcolor{blue}{\tt {\color{white}001}} & {\tt HD} \\ \hline
      Aquatic plants and sea-grass & \cellcolor{green}{\tt 010} & {\tt PF} \\ \hline
      Wrecks or ruins & \cellcolor{sky}{\tt 011} & {\tt WR} \\ \hline
      Robots (AUVs/ROVs/instruments) & \cellcolor{red}{\tt 100} & {\tt RO} \\ \hline
      Reefs and invertebrates & \cellcolor{pink}{\tt 101} & {\tt RI} \\ \hline
      Fish and vertebrates & \cellcolor{yellow}{\tt 110} & {\tt FV} \\ \hline
      Sea-floor and rocks  & \cellcolor{white}{\tt {111}} & {\tt SR} \\
      \hline
    \end{tabular}
    \label{tab:my_label}
\end{table}

\begin{figure}[hb]
    \centering
    \begin{subfigure}{0.5\textwidth} 
    \centering
        \includegraphics[width=0.75\linewidth]{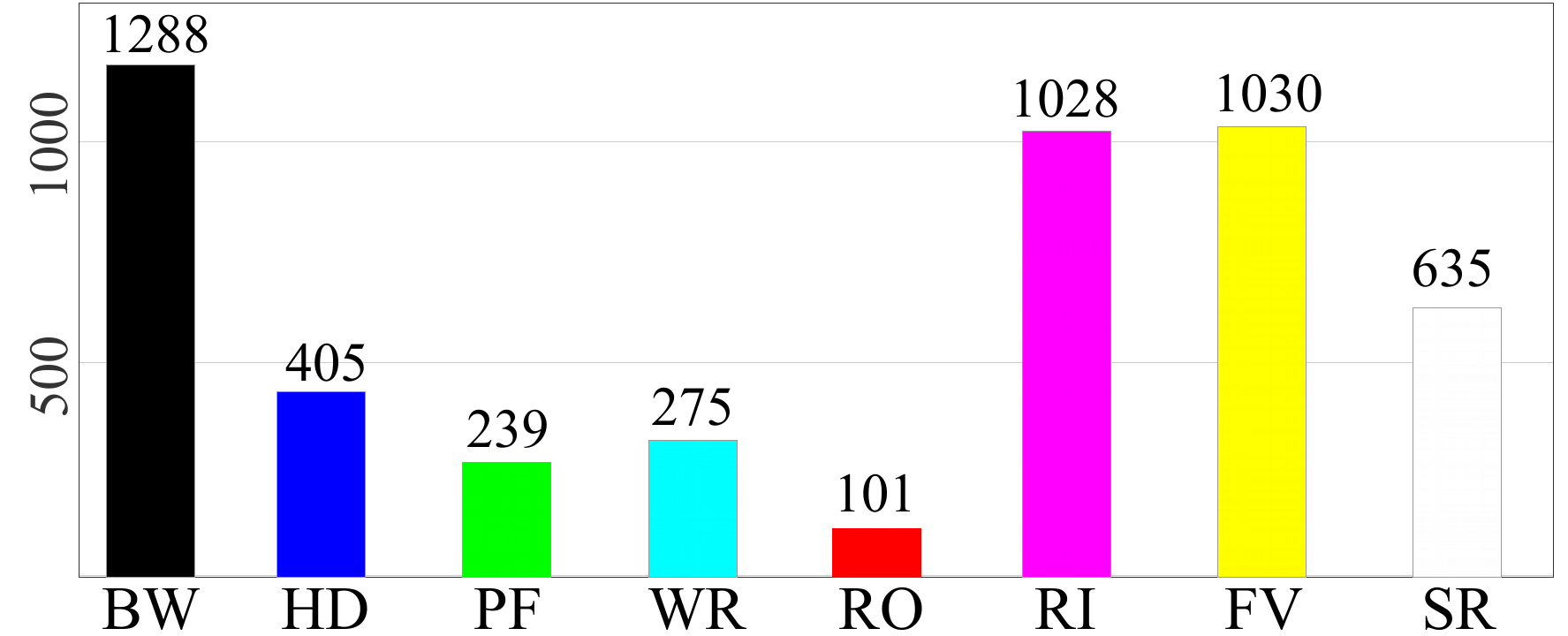}
        \caption{The number of images containing each object category.}
        \label{data2}
    \end{subfigure}
    \vspace{2mm}
    
    \begin{subfigure}{0.5\textwidth}
    \centering
        \includegraphics[width=0.65\linewidth]{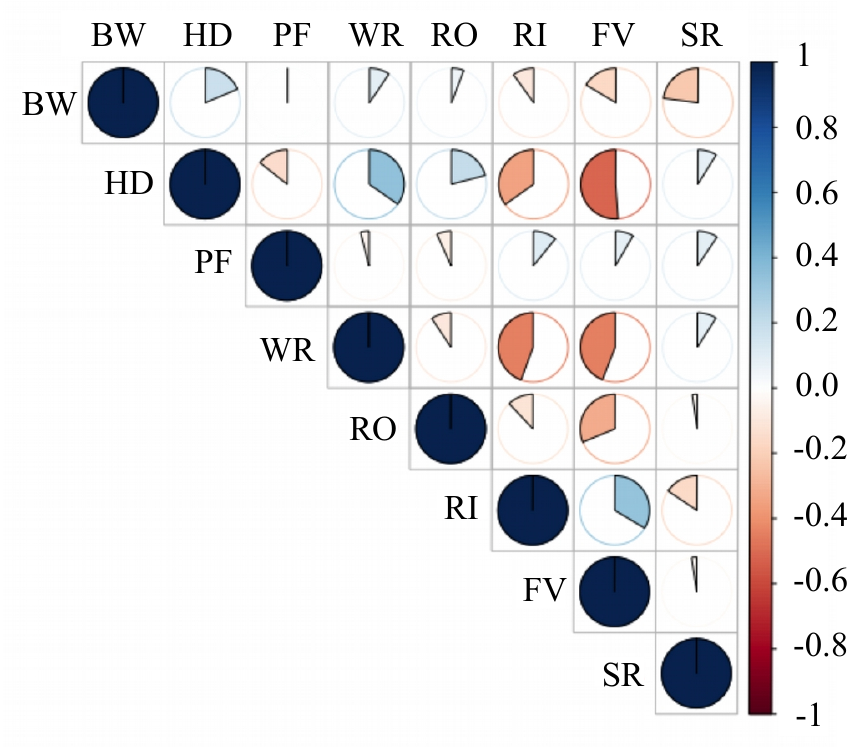}
        \caption{Pairwise correlations of the object categories' occurrences.}
        \label{data5}
    \end{subfigure}
    \vspace{2mm}
    
    \begin{subfigure}{0.5\textwidth}
    \centering
        \includegraphics[width=0.65\linewidth]{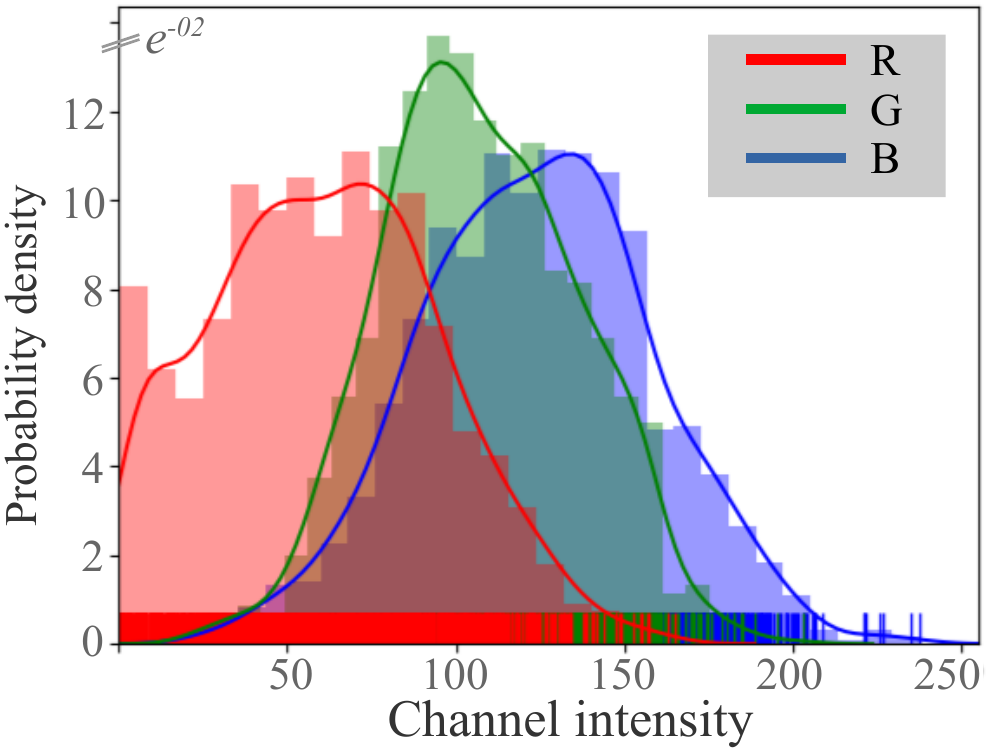}
        \caption{Distributions of averaged pixel intensity values.}
        \label{data3}
    \end{subfigure}
    
    \caption{Statistics concerning various object categories and image intensity values in the SUIM dataset.}
    \label{fig:data_stat}
\end{figure}

The SUIM dataset has $1525$ RGB images for training and validation; another $110$ test images are provided for benchmark evaluation of semantic segmentation models. The images are of various spatial resolutions, \eg, $1906 \times 1080$, 
$1280 \times 720$, $640 \times 480$, and $256 \times 256$. These images are carefully chosen from a large pool of samples collected during oceanic explorations and human-robot cooperative experiments in several locations of various water types. We also utilize a few images from large-scale datasets named EUVP~\cite{islam2019fast}, USR-248~\cite{islam2019srdrm}, and UFO-120~\cite{islam2020sesr}, which we previously proposed for underwater image enhancement and super-resolution problems. The images are chosen to accommodate a diverse set of natural underwater scenes and various setups for human-robot collaborative experiments. Figure~\ref{fig:data_stat} demonstrates the population of each object category, their pairwise correlations, and the distributions of RGB channel intensity values in the SUIM dataset.

\begin{figure*}[ht]
    \centering
    \centering
    \includegraphics[width=0.99\linewidth]{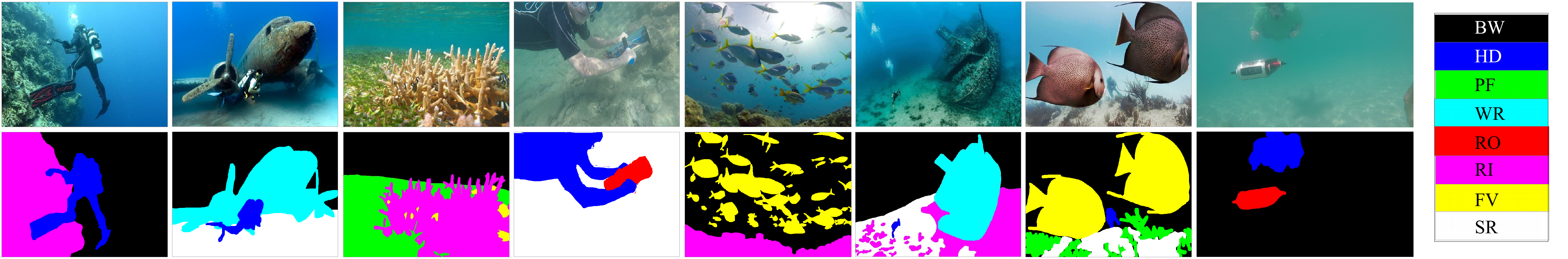} 
    \caption{A few sample images and corresponding pixel-annotations are shown on the top, and bottom row, respectively.}
    \label{fig:data}
\end{figure*}

All images of the SUIM dataset are pixel-annotated by seven human participants; a few samples are shown in Figure~\ref{fig:data}. We followed the guidelines discussed in~\cite{ConfRef1} and~\cite{ConfRef2} for classifying potentially confusing objects of interest such as plants/reefs, vertebrates/invertebrates, etc.

%% file: src/rel.tex
\section{Use Cases and Related Work}\label{related}
\subsection{Semantic Segmentation}
The learning-based semantic segmentation methodologies have made remarkable progress over the last decade due to the advent of powerful deep models~\cite{garcia2017review} and large-scale annotated datasets (of mostly terrestrial images~\cite{oquab2014learning,lin2014microsoft} and videos~\cite{perazzi2016benchmark}).
The forerunner approaches are based on fully convolutional networks (FCNs)~\cite{long2015fully} that utilize the seminal CNN-based models (\eg, VGG~\cite{simonyan2014very}, GoogLeNet~\cite{szegedy2015going}, ResNet~\cite{he2016deep}) for hierarchical feature extraction. The \emph{encoded} semantic information is then exploited by a \emph{decoder} network that learns to classify each pixel; it gradually up-samples the low-dimensional features by a series of deconvolution layers~\cite{zeiler2010deconvolutional} and eventually generates the pixel-wise labels. 
More effective learning pipelines are later proposed for integrating global contextual information and instance awareness. For instance, SegNet architectures~\cite{badrinarayanan2015segnet} accommodate the mapping of deep encoder layers' output features into input dimensions rather than performing ad hoc up-sampling. Moreover, UNet architectures~\cite{ronneberger2015u} reuse each encoder layers' output by skip-connections to mirrored decoder layers, which significantly improves performance. 
Several other contemporary approaches incorporate capabilities such as global feature fusion~\cite{liu2015parsenet,zhao2017pyramid}, spatio-temporal learning~\cite{visin2016reseg}, multi-scale learning~\cite{roy2016multi}, etc. The notion of \emph{atrous convolution}~\cite{chen2017deeplab} (aka dilated convolution~\cite{yu2015multi}) and conditional random field (CRF)-based post-processing stages (introduced in the DeepLab models~\cite{chen2017deeplab,chen2017rethinking}) further ensure multi-scale
context awareness and fine-grained localization of object boundaries.

Despite the advancements, semantic segmentation of underwater imagery is considerably less studied. 
Moreover, the existing solutions for terrestrial imagery are not directly applicable because the object categories and image statistics are entirely different. A unique set of underwater image distortion artifacts and the unavailability of large-scale annotated datasets further influence a significant lack of research attempts. Several important contributions address the problems of coral-reef classification and segmentation~\cite{alonso2019coralseg,beijbom2012automated}, coral-reefs' coverage estimation~\cite{beijbom2012automated}, fish detection and segmentation~\cite{VAIME,ravanbakhsh2015automated,chuang2011automatic}, etc. However, these application-specific models are not feasible for general use in robotic applications. Other classical approaches use fuzzy C-means clustering and stochastic optimization methods for foreground segmentation~\cite{li2016mapreduce,arain2019improving,padmavathi2010non}. Such class-agnostic approaches group image regions by evaluating local salient features~\cite{zhu2017underwater,ye2009objective}; hence, these cannot be generalized for multi-object semantic segmentation.

Nevertheless, a few recent work~\cite{zhou2019underwater,islam2020sesr} explored the performance of contemporary deep CNN-based semantic segmentation models such as VGG-based encoder-decoders~\cite{simonyan2014very,garcia2017review}, UNet~\cite{ronneberger2015u}, and SegNet~\cite{badrinarayanan2015segnet} for underwater imagery.     
Although they report inspiring results, they only consider sea-grass, sand, and rock as object categories. Moreover, performance evaluations of many important SOTA models are still unexplored. We attempt to address these limitations by considering a more comprehensive set of object categories in the SUIM dataset, and provide a thorough benchmark evaluation of the SOTA semantic segmentation models (see Section~\ref{sec:benchmark}).

\subsection{Visual Attention Modeling and Servoing}
\begin{figure}[ht]
    \centering
    \includegraphics[width=0.99\linewidth]{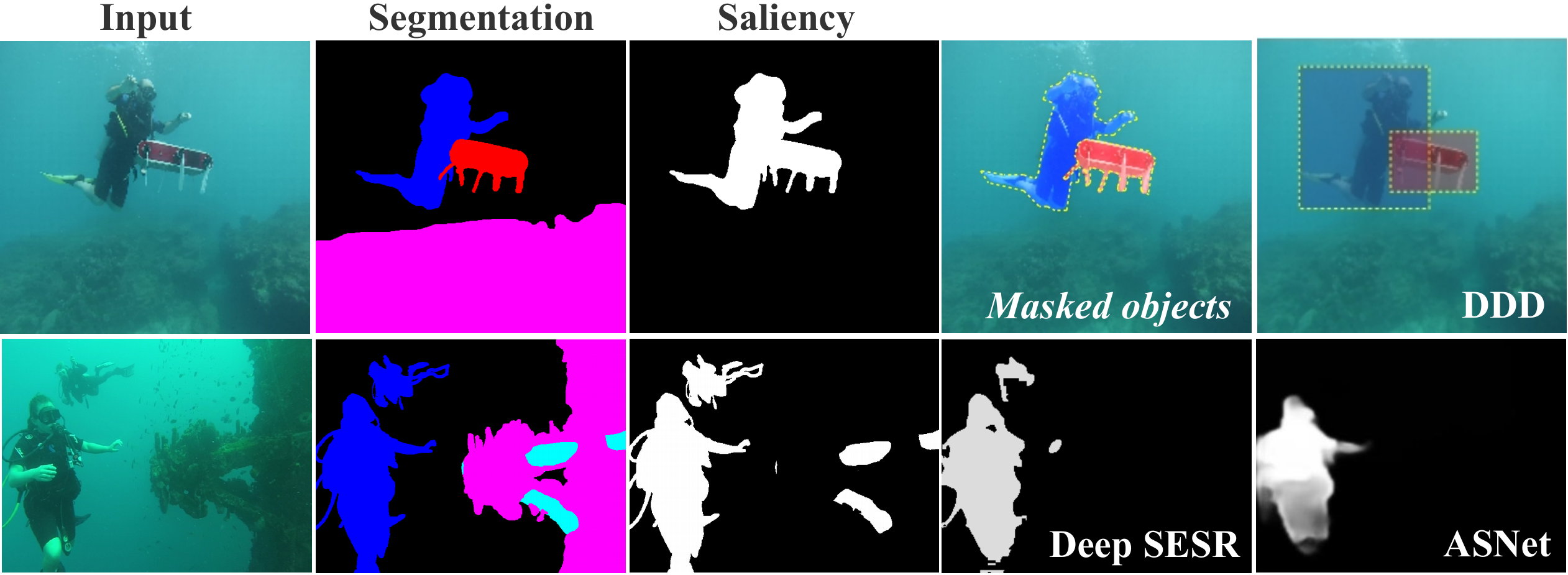}
    \caption{Generation of semantic saliency maps from segmentation masks: the intensities of {\tt HD}, {\tt RO}, {\tt FV}, and {\tt WR} pixels are set to $1.0$, and the rest are set to $0.0$. In comparison: output of a CNN-based object detector named DDD~\cite{islam2018towards} (top); and two class-agnostic saliency predictors named Deep SESR~\cite{islam2020sesr} and ASNet~\cite{wang2018salient} (bottom).}
    \label{fig:rel}
\end{figure}

`Where to look'- is an important open problem for autonomous underwater exploration and surverying~\cite{shkurti2012multi,bingham2010robotic}. In particular, the most essential capability of visually-guided AUVs is to analyze image-based features for modeling relative attention~\cite{girdhar2014autonomous,kaeli2014visual} of various regions of interest (RoIs). Such visual attention-based cues are eventually exploited to make important navigational and other operational decisions. The classical approaches utilize features such as luminance, color, texture, and often depth information to extract salient features for enhanced object detection or template identification~\cite{maldonado2016robotic,zhang2016underwater,islam2017mixed}. 
In recent years, the standard one-shot object detection models based on large-scale supervised learning are effectively applied for vision-based tracking and following~\cite{shkurti2017underwater,islam2018understanding}. 
These general-purpose object detectors are then coupled with application-specific bounding box (BBox)-reactive controllers for visual servoing~\cite{shkurti2017underwater,islam2018towards}. 
Nevertheless, semantic segmentation provides pixel-level detection accuracy and tighter object boundaries than a BBox (see Figure~\ref{fig:rel}), which are useful for more robust tracking.

% towards object detection and tracking
Another genre of approaches for visual attention modeling focuses on spatial \emph{saliency} computation, \ie, predicting relevance/importance of each pixel in the image~\cite{islam2020sesr,wang2018salient}. 
The salient image regions can be exploited for the detection and tracking of specific (known) objects, or for finding new objects of interest in exploration tasks~\cite{girdhar2014autonomous}. For human-robot collaborative applications, in particular, an AUV needs to keep its companion humans/robots within the field-of-view, and pay attention to other objects in the scene, \eg, fish, reefs, wrecks, etc. 
A particular instance of such semantic saliency computation is demonstrated in Figure~\ref{fig:rel}. While a class-agnostic saliency map provides interesting foreground regions, the semantic saliency map embeds additional information about the spatial distribution and interaction among the objects in the scene. 
Moreover, this semantic knowledge is potentially useful for learning spatio-temporal attention modeling and visual question answering~\cite{yu2017multi,bazzani2016recurrent}, \ie, finding image regions that are relevant to a given query. These exciting research problems have not been explored in depth for underwater robotic applications.

%% file: src/suimnet.tex
\section{The SUIM-Net Model}
\subsection{Network Architecture}
\begin{figure}[ht]
    \centering
    \begin{subfigure}{0.5\textwidth} 
    \centering
        \includegraphics[width=0.98\linewidth]{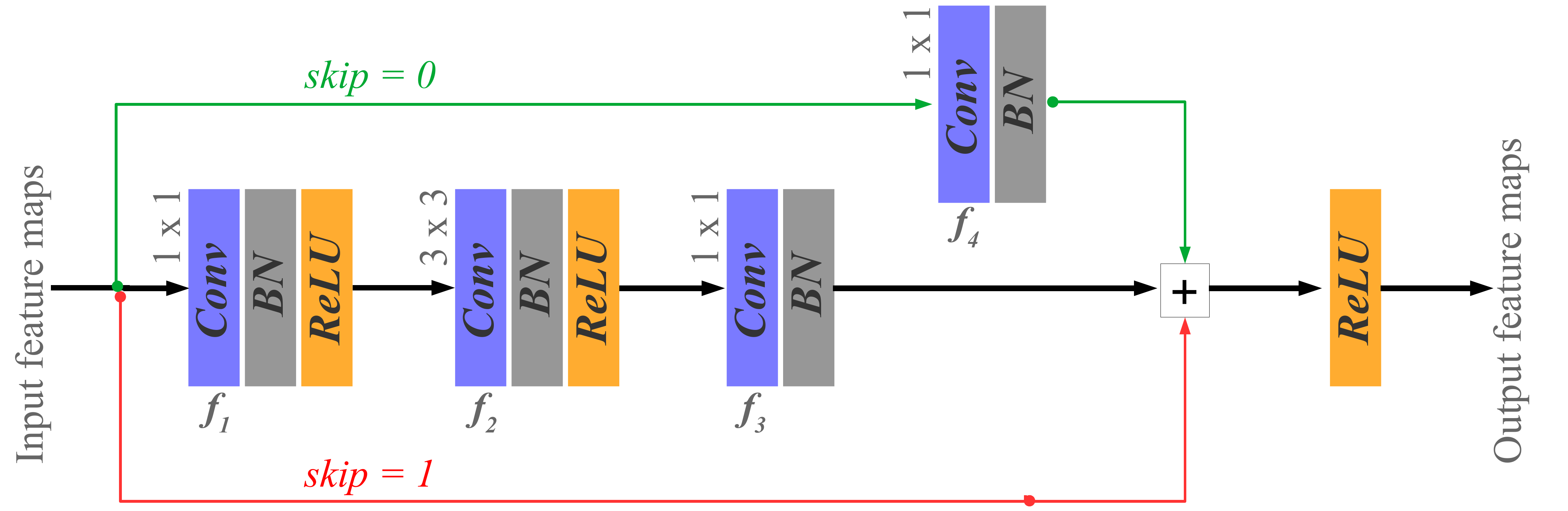}
        \caption{Architecture of an RSB: the skip-connection can be either fed from an intermediate {\tt conv} layer (by setting \textit{skip=0}) or from the input (by setting \textit{skip=1}) for local residual learning.}
        \label{fig:model_a}
    \end{subfigure}
    \vspace{2mm}
    
    \begin{subfigure}{0.5\textwidth} 
    \centering
    \includegraphics[width=0.99\linewidth]{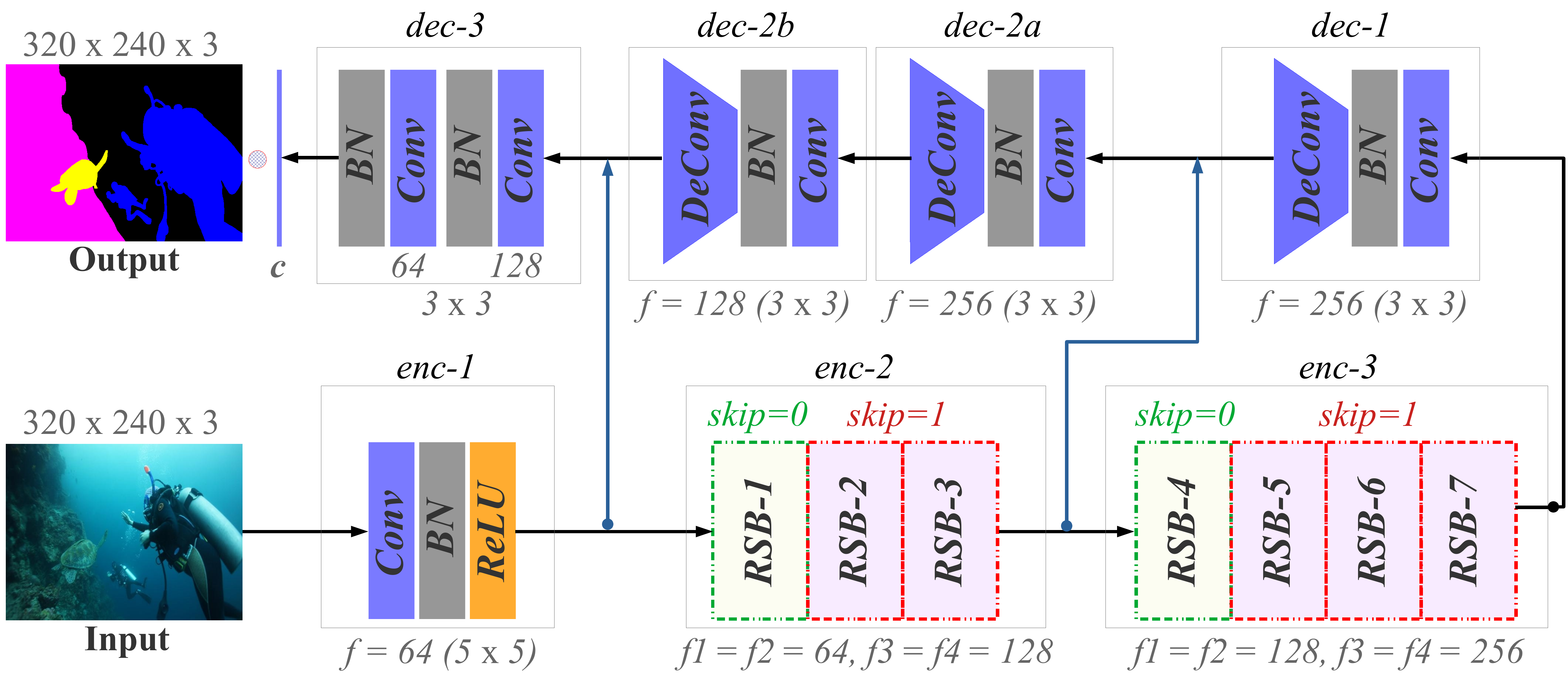}
    \caption{The end-to-end architecture of SUIM-Net$_{RSB}$: three composite layers of encoding is performed by a total of seven RSBs, followed by three decoder blocks with mirrored skip-connections.}
    \label{fig:model_b}
    \end{subfigure}
    \vspace{2mm}
    
    \begin{subfigure}{0.5\textwidth} 
    \centering
    \includegraphics[width=0.99\linewidth]{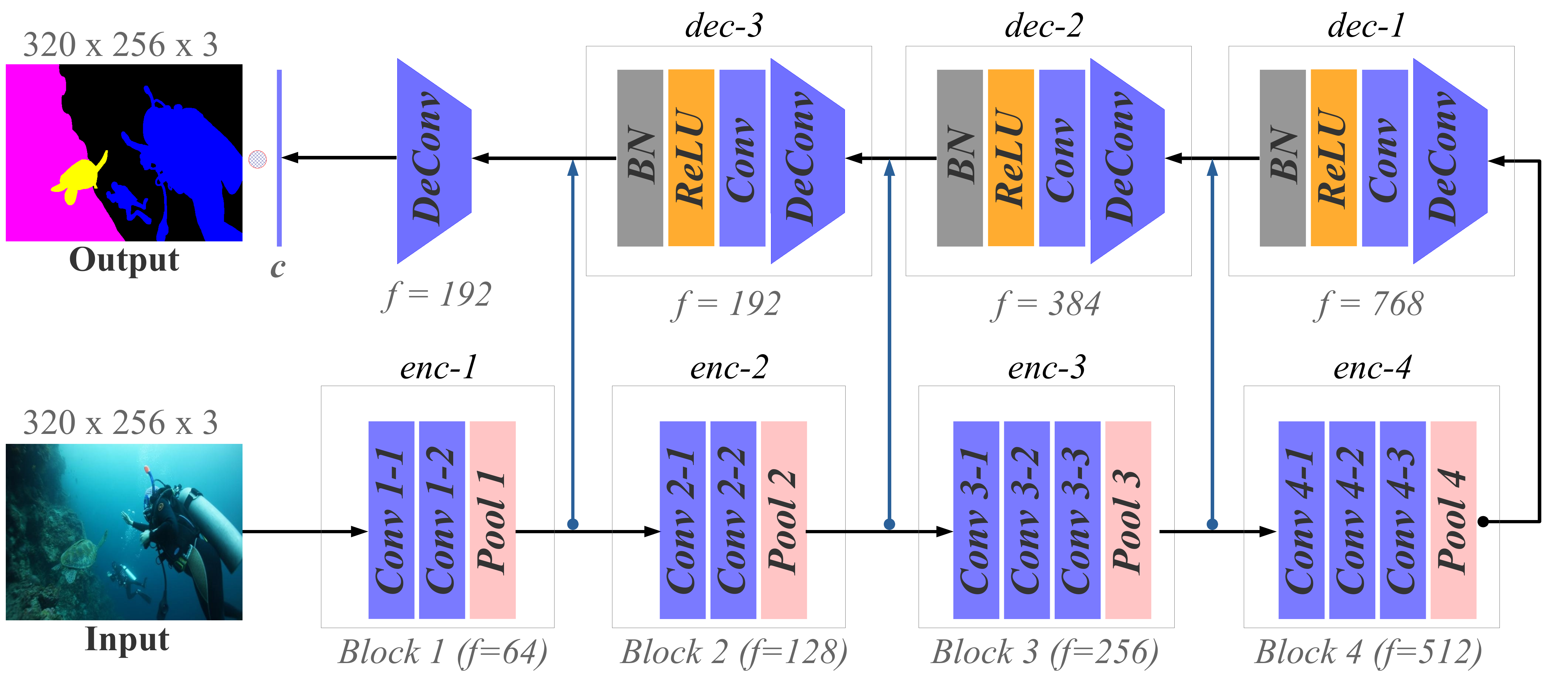}
    \caption{The end-to-end architecture of SUIM-Net$_{VGG}$: first four blocks of a pre-trained VGG-16~\cite{simonyan2014very} model are used for encoding, followed by three mirrored decoder blocks and a {\tt deconv} layer.}
    \label{fig:model_c}
    \end{subfigure}
    
    \caption{Detailed specification of the SUIM-Net model.}
    \label{fig:model}
\end{figure}

The proposed SUIM-Net model incorporates a fully convolutional encoder-decoder architecture with skip connections between mirrored composite layers. As shown in Figure~\ref{fig:model_a}, the base model embodies residual learning~\cite{he2016deep} with an optional skip layer named RSB (residual skip block). Each RSB consists of three convolutional ({\tt conv}) layers, each followed by Batch Normalization ({\tt BN})~\cite{ioffe2015batch} and {\tt ReLU} non-linearity~\cite{nair2010rectified}. As Figure~\ref{fig:model_b} shows, two sets of RSBs are used sequentially in the second and third encoder layers; the number of filters, feature dimensions, and other parameters are marked in the figure. The encoder network extracts $256$ feature maps from input RGB images; the encoded feature maps are then exploited by three sequential decoder layers. Each decoder layer consists of a {\tt conv} layer that receives skip-connections from their respective conjugate encoder layer; it is followed by {\tt BN} and a de-convolutional ({\tt deconv}) layer~\cite{zeiler2010deconvolutional} for spatial up-sampling. The final {\tt conv} layer subsequently generates the per-channel binary pixel labels for each object category, which can be post-processed for visualizing in the RGB space.

As evident from Figure~\ref{fig:model_b}, we try to combine the benefits of skip-connections~\cite{ronneberger2015u} and residual learning~\cite{he2016deep,badrinarayanan2015segnet} in the \emph{SUIM-Net$_{RSB}$} model. Our motive for such design is to ensure real-time inference while achieving a reasonable segmentation performance. 
On the other hand, we focus more on improved performance in the \emph{SUIM-Net$_{VGG}$} model, which utilizes $12$ encoding layers of a pre-trained VGG-16 network~\cite{simonyan2014very}; detailed architecture is illustrated in Figure~\ref{fig:model_c}. 
We present the experimental analysis of SUIM-Net and other SOTA models in the following sections. 

%\begin{table}[b]
%\centering
%\caption{The input dimensions and training parameters for all the models in comparison. [$e$: number of epochs; $s$: steps per epoch; $b$: batch size]}
%\scriptsize
%\vspace{-1mm}
%\begin{tabular}{l||c|c|c}
%  \hline
%  \textbf{Model} & Input dimension & $e \odot s$ & $b$  \\ \hline \hline
%  FCN8$_{CNN}$ &$320\times240\times3$   &  $50 \odot 4000$   &  $2$ \\
%  FCN8$_{VGG}$ &$320\times240\times3$   &  $40 \odot 4000$   &  $2$ \\ \hline
%  SegNet$_{CNN}$ & $320\times256\times3$  &  $40 \odot 4000$    & $8$  \\
%  SegNet$_{ResNet}$ & $320\times256\times3$  &  $50 \odot 5000$    & $4$  \\  \hline
%  UNet$_{GREY}$ & $320 \times 240 \times 1$  &   $20 \odot 4000$   & $4$  \\
%  UNet$_{RGB}$ & $320\times240\times3$  &  $30 \odot 5000$    & $2$  \\ \hline
%  PSPNet$_{MobileNet}$ & $384\times384\times3$  &  $60 \odot 5000$   & $2$   \\ \hline
%  DeepLab$_{V3}$ & $320\times320\times3$  & $50 \odot 4000$    &  $2$ \\ \hline
%  SUIM-Net & $320\times240\times3$  &  $45 \odot 5000$   & $4$   \\
%  \hline
%\end{tabular}
%\label{tab:train}
%\end{table}%

\subsection{Training Pipeline and Implementation Details}
We formulate the problem as learning a mapping from input domain $X$ (of natural underwater images) to its target semantic labeling $Y$ in RGB space. %We consider eight object categories of the SUIM dataset and its paired annotated data to learn the underlying function $G: X \rightarrow Y$. 
The end-to-end training is supervised by the standard cross-entropy loss~\cite{zhang2018generalized,badrinarayanan2015segnet} which evaluates the discrepancy between predicted and ground truth pixel labels. We use TensorFlow libraries~\cite{abadi2016tensorflow} to implement the optimization pipeline; a Linux host with one Nvidia\texttrademark{ }GTX 1080 graphics card is used for training. Adam optimizer~\cite{kingma2014adam} is used for the global iterative learning with a rate of $10^{-4}$ and a momentum of $0.5$. Moreover, we apply several image transformations for data augmentation during training, which are specified in Appendix I.

%% file: src/benchmark.tex
\begin{table*}[ht]
\centering
\caption{Quantitative performance comparison for semantic segmentation and saliency prediction: scores are shown as $\textit{mean} \pm \sqrt{\textit{variance}}$; the {\color{red}best score} and {\color{blue}next top three scores} for each comparison are colored red and blue, respectively.}
\footnotesize
%\scriptsize
\vspace{-1mm}
\begin{tabular}{c|l||c|c|c|c|c|c||c}
  \hline
  & \textbf{Model} & {\tt HD} & {\tt WR} & {\tt RO} & {\tt RI} & {\tt FV} & \textbf{Combined} & Saliency Pred.  \\ \hline \hline
  % change this number 7 when adding/deleting rows
  \parbox[t]{2mm}{\multirow{8}{*}{\rot{$\mathcal{F}$ ($\rightarrow$)}}} & FCN8$_{CNN}$ & $76.34\pm 2.24$ & $70.24 \pm 2.26$  & $39.83 \pm 3.87$  & $61.65 \pm 2.36$ & $76.24 \pm 1.87$ & $64.86 \pm 2.52$ & $75.62 \pm 1.79$ \\ %\hline
  
  & FCN8$_{VGG}$ & {\color{blue}$89.10 \pm 1.50$} & {\color{blue}$82.03 \pm 1.94$} & {\color{blue}$74.01 \pm 3.23$} & {\color{blue}$79.19 \pm 2.27$} & {\color{blue}$90.46 \pm 1.18$}
 & {\color{blue}$82.96 \pm 2.02$} &  {\color{blue}$89.63 \pm 1.24$} \\ %\cline{2-9} 
  
  & SegNet$_{CNN}$ & $59.60 \pm 2.02$ & $41.60 \pm 1.65$  & $31.77 \pm 3.03$ & $41.88 \pm 2.66$ & $60.08 \pm 1.91$ & $46.97 \pm 2.25$ & $56.96 \pm 1.58$  \\ %\hline
  
  & SegNet$_{ResNet}$ & $80.52 \pm 3.26$  & {$77.65 \pm 3.15$}  & $62.45 \pm 3.90$ & {\color{blue}$82.30 \pm 1.96$} & {\color{blue}$91.47 \pm 1.01$} & $76.88 \pm 2.66$ & {\color{blue}$86.88 \pm 1.83$}  \\ %\hline

  & UNet$_{GRAY}$ &  $85.47 \pm 2.21$ & {\color{blue}$79.77 \pm 2.01$} & $60.95 \pm 3.31$ & $69.95 \pm 2.57$ & $84.47 \pm 1.39$ & $75.12 \pm 2.30$ & $83.96 \pm 1.40$ \\ %\hline  \\ 
  
  & UNet$_{RGB}$ & {\color{blue}$89.60 \pm 1.84$} & {\color{red}$86.17 \pm 1.73$}  & {$68.87 \pm 3.30$}  & {\color{blue}$79.24 \pm 2.70$} & {\color{blue}$91.35 \pm 1.14$} & {\color{blue}$83.05 \pm 2.14$} & {\color{blue}$89.99 \pm 1.29$}  \\ %\hline

  & PSPNet$_{MobileNet}$ &  {$80.21 \pm 1.19$} & $70.94 \pm 1.61$  & {$72.04 \pm 2.21$} & $72.65 \pm 1.62$ & $79.19 \pm 1.74$ & $76.01 \pm 1.67$ & $78.42 \pm 1.59$ \\ %\hline

  & DeepLab$_{V3}$ &   {\color{blue}$89.68 \pm 2.09$} & {$77.73 \pm 2.18$}  & {\color{blue}$72.72  \pm 3.35$}   & {$78.28 \pm 2.70$}  & {$87.95 \pm 1.59$}  & {\color{blue}$81.27 \pm 2.30$} & {$85.94 \pm 1.72$} \\ %\hline
  
  & \textbf{SUIM-Net$_{RSB}$} & {$89.04 \pm 1.31$} & $65.37 \pm 2.22$ & {\color{blue}$74.18 \pm 2.11$} & $71.92 \pm 1.80$ & $84.36 \pm 1.37$ & {$78.86 \pm 1.79$} & $81.36 \pm 1.72$ \\ %\hline  \\ 
  
  & \textbf{SUIM-Net$_{VGG}$} & {\color{red}$93.56 \pm 0.98$} & {\color{blue}$86.02 \pm 1.03$} & {\color{red}$78.06 \pm 1.50$} & 
  {\color{red}$83.49 \pm 1.39$} & 
  {\color{red}$93.73 \pm 0.87$} & {\color{red}$86.97 \pm 1.15$} & 
  {\color{red}$91.91 \pm 0.85$} \\ %\hline  \\ 
  \hline 
  
    % change this number 7 when adding/deleting rows
  \parbox[t]{2mm}{\multirow{8}{*}{\rot{$mIOU$ ($\rightarrow$)}}} 
  & FCN8$_{CNN}$ & $67.27 \pm 2.50$  &$81.64 \pm 2.16$   & $36.44 \pm 3.67$ &$78.72 \pm 2.50$  & $70.25 \pm 2.28$  
  & $66.86 \pm 2.62$ & $75.63 \pm 1.89$ \\ %\hline

  & FCN8$_{VGG}$ & {$79.86 \pm 1.50$} & {\color{blue}$85.77 \pm 2.09$} & {$65.05 \pm 3.00$} & {\color{blue}$85.23 \pm 2.07$} & {\color{blue}$81.18 \pm 1.46$} & {\color{blue}$79.42 \pm 2.02$} & {\color{blue}$85.22 \pm 1.24$} \\ %\cline{2-9} 
 
  & SegNet$_{CNN}$ & $62.76 \pm 2.35$ & $66.75 \pm 2.57$  & $36.63 \pm 3.12$ & $63.46 \pm 3.18$ & $62.48 \pm 2.32$ & $58.42 \pm 2.71$ & $65.90 \pm 2.12$ \\ %\hline
  
  & SegNet$_{ResNet}$ & $74.00 \pm 2.88$ & $82.68 \pm 2.94$ & $58.63 \pm 3.61$ & {\color{red}$89.61 \pm 1.15$} & {\color{blue}$82.96 \pm 1.38$} & $77.58 \pm 2.39$ & {$83.09 \pm 1.96$} \\ %\hline
  
  & UNet$_{GRAY}$ &  {$78.33 \pm 2.34$} & {$85.14\pm 2.14$} & $57.25 \pm 3.00$ & $79.96 \pm 2.55$ & $78.00 \pm 1.90$ &  $75.74 \pm 2.38$ & $82.77 \pm 1.59$ \\ %\hline 

  & UNet$_{RGB}$ & {\color{blue}$81.17 \pm 2.02$} &  {\color{blue}$87.54 \pm 2.00$} & $62.07 \pm 3.12$  & {$83.69 \pm 2.58$} & {\color{red}$83.83 \pm 1.47$} & {\color{blue}$79.66 \pm 2.24$} & {\color{blue}$85.85 \pm 1.54$}  \\ %\hline
  
  & PSPNet$_{MobileNet}$ & $75.76 \pm 1.47$ & {\color{blue}$86.82 \pm 1.26$} & {\color{red}$72.66 \pm 1.47$}  & {\color{blue}$85.16 \pm 1.65$} & $74.67 \pm 1.90$ & $77.41 \pm 1.56$ & $80.87 \pm 1.56$ \\ %\hline

  & DeepLab$_{V3}$ & {\color{blue}$80.78 \pm 2.07$}  & {$85.17 \pm 2.08$}  & {\color{blue}$66.03 \pm 3.16$}  & {$83.96 \pm 2.52$} & {$79.62 \pm 1.85$}  & {\color{blue}$79.10 \pm 2.34$} & {\color{blue}$83.55 \pm 1.65$} \\ %\hline
  
  & \textbf{SUIM-Net$_{RSB}$} & {\color{blue}$81.12 \pm 1.76$}  & $80.68 \pm 1.74$ & {\color{blue}$65.79 \pm 2.10$} & {$84.90 \pm 1.77$} & $76.81 \pm 1.82$ & {$77.77 \pm 1.64$}& $80.86 \pm 1.64$  \\ %\hline  \\
  
  & \textbf{SUIM-Net$_{VGG}$} & {\color{red}$85.09 \pm 1.45$}  & {\color{red}$89.90 \pm 1.29$} & {\color{blue}$72.49 \pm 1.61$} & {\color{blue}$89.51 \pm 1.25$} & 
  {\color{blue}$83.78 \pm 1.55$} & {\color{red}$84.14 \pm 1.43$} & 
  {\color{red}$87.67 \pm 1.24$} \\ %\hline  \\
  \hline
  
\end{tabular}
\label{tab:quan_se}
%\vspace{-1mm}
\end{table*}

\section{Benchmark Evaluation}~\label{sec:benchmark}
We consider the following SOTA models for performance evaluation on the SUIM dataset: 
\textit{i)} FCN8~\cite{long2015fully} with two variants of base model: vanilla CNN (FCN8$_{CNN}$) and VGG-16 (FCN8$_{VGG}$), 
\textit{ii)}  SegNet~\cite{badrinarayanan2015segnet} with two variants of base model: vanilla CNN (SegNet$_{CNN}$) and ResNet-50 (SegNet$_{ResNet}$), \textit{iii)}  UNet~\cite{ronneberger2015u} with two variants of input: grayscale images (UNet$_{GRAY}$) and RGB images (UNet$_{RGB}$), iv) pyramid scene parsing network~\cite{zhao2017pyramid} with MobileNet~\cite{howard2017mobilenets} as the base model (PSPNet$_{MobileNet}$), and \textit{v)} DeepLab$_{V3}$~\cite{chen2017rethinking}. We use TensorFlow implementations of all these models and train them on the SUIM datasest using the same hardware setup (as of SUIM-Net); further information can be found in their source repositories which are provided in Appendix II.  

We mentioned various use cases of the SUIM dataset for semantic segmentation and saliency prediction in Section~\ref{related}. In our evaluation, we conduct performance comparison of the SOTA models for the following two training configurations: 
\begin{itemize}
    \item Semantic segmentation with the five major object categories (see Table~\ref{tab:my_label}): {\tt HD}, {\tt WR}, {\tt RO}, {\tt RI}, and {\tt FV}; the rest are considered as background, \ie, {\tt BW}={\tt PF}={\tt SR}={\tt (000)\textsubscript{RGB}}. Each model is configured for five channels of output, one for each category. The predicted separate pixel masks are combined to RGB masks for visualization.

    \item Single-channel saliency prediction: the ground truth intensities of {\tt HD}, {\tt RO}, {\tt FV}, and {\tt WR} pixels are set to $1.0$, and the rest are set to $0.0$. The output is thresholded and visualized as binary images.    
\end{itemize}
Detailed performance analysis for these two setups is presented in the following sections.

\subsection{Evaluation Criteria}
We compare the performance of all the models based on standard metrics that evaluate region similarity and contour accuracy~\cite{garcia2017review,perazzi2016benchmark}. The region similarity metric quantifies the correctness of predicted pixel labels compared to ground truth by using the notion of `dice coefficient' aka $\mathcal{F}$ score. It is calculated using the precision ($\mathcal{P}$) and recall ($\mathcal{R}$) as $\mathcal{F}=\frac{2 \times \mathcal{P} \times \mathcal{R}}{\mathcal{P}+\mathcal{R}}$. On the other hand, contour accuracy represents the object boundary localization performance; it is quantified by the mean IOU (intersection over union) scores, where $IOU = \frac{Area\text{ }of\text{ }overlap}{Area\text{ }of\text{ }union}$.

\begin{figure*}[t]
    \centering
    \centering
    \includegraphics[width=0.99\linewidth]{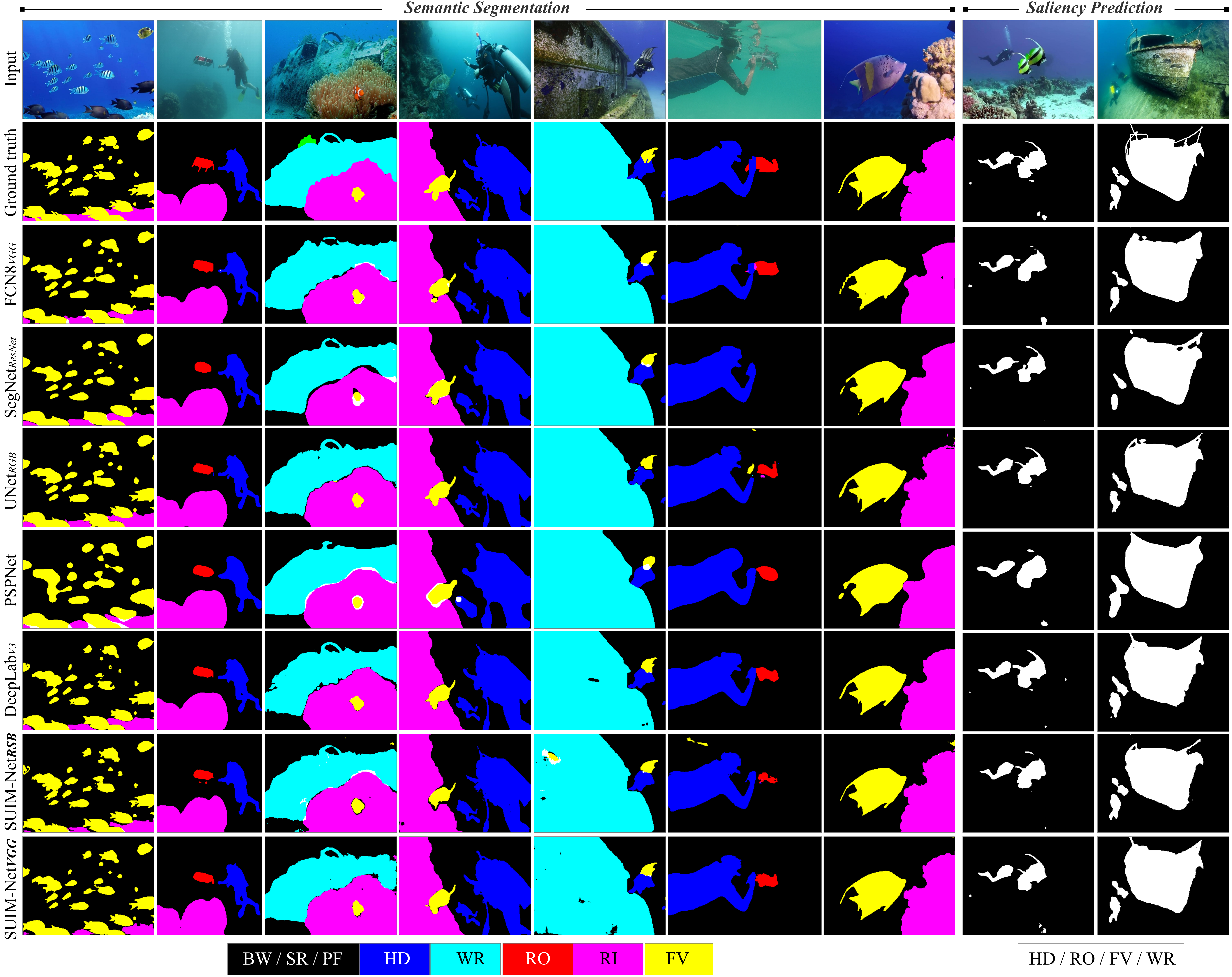}
    \caption{A few qualitative comparison for the experiment of Table~\ref{tab:quan_se}: (left) semantic segmentation with {{\tt HD}}, {{\tt WR}}, {{\tt RO}}, {{\tt RI}}, and {{\tt FV}} as object categories; (right) saliency prediction with {\tt HD}={\tt RO}={\tt FV}={\tt WR}={\tt 1} and {\tt RI}={\tt PF}={\tt SR}={\tt BW}={\tt 0}. Results for the top  performing models are shown; best viewed digitally by zoom for details.}
    \label{fig:qual}
\end{figure*}

\subsection{Quantitative and Qualitative Analysis}
We present the quantitative results in Table~\ref{tab:quan_se}. It compares the $\mathcal{F}$ and \textit{mIOU} scores for semantic segmentation of each object category; it also compares the respective scores for saliency prediction. The results suggest that UNet$_{RGB}$, FCN8$_{VGG}$, and DeepLab$_{V3}$ generally perform better than other models. In particular, they achieve the top three $\mathcal{F}$ and \textit{mIOU} scores for both semantic segmentation and saliency prediction. SegNet$_{ResNet}$ and PSPNet$_{MobileNet}$ also provide competitive results; however, their performances are slightly inconsistent over various object categories. Moreover, significantly better scores of SegNet$_{ResNet}$ (FCN8$_{VGG}$) over SegNet$_{CNN}$ (FCN8$_{CNN}$) validate the benefits of using a powerful feature extractor. As Table~\ref{tab:usability} shows, SegNet$_{ResNet}$ (FCN8$_{VGG}$) has about twice (five times) the number of network parameters than SegNet$_{CNN}$ (FCN8$_{CNN}$). On the other hand, consistently better performance of UNet$_{RGB}$ over UNet$_{GRAY}$ validates the utility of learning on RGB image space (rather than using grayscale images as input).

\begin{table}[ht]
\centering
\caption{Comparison for the input resolution, number of model parameters, and inference rates (averaged end-to-end processing times on a single Nvidia\texttrademark~GTX 1080 GPU).}
%\scriptsize
%\vspace{-1mm}
\begin{tabular}{l||c|r|r}
  \hline
  \textbf{Model} & Resolution & \# of Parameters & Frame rate  \\ \hline \hline
  FCN8$_{CNN}$ & $320\times240$ & $69.744$ M & $17.11$ FPS \\
  FCN8$_{VGG}$& $320\times240$ & $134.286$ M & $8.79$ FPS \\
  SegNet$_{CNN}$& $320\times256$ & $2.845$ M & $17.52$ FPS \\
  SegNet$_{ResNet}$& $320\times256$ & $15.012$ M & $10.86$ FPS \\
  UNet$_{GRAY}$& $320\times240$ &$31.032$ M & $20.13$ FPS \\
  UNet$_{RGB}$& $320\times240$ & $31.033$ M & $19.98$ FPS \\
  PSPNet$_{MobileNet}$& $384\times384$ & $63.967$ M & $6.65$ FPS  \\
  DeepLab$_{V3}$& $320\times320$ &$41.254$ M  & $16.00$ FPS \\
  \textbf{SUIM-Net$_{RSB}$}& $320\times240$ & $\mathbf{3.864}$ \textbf{M} & {$\mathbf{28.65}$} \textbf{FPS} \\
  \textbf{SUIM-Net$_{VGG}$}& $320\times256$ & $\mathbf{12.219}$ \textbf{M} & {$\mathbf{22.46}$} \textbf{FPS} \\
  \hline
\end{tabular}
\label{tab:usability}
\end{table}%

%\begin{table}[h]
%\centering
%\caption{Comparison for the number of parameters and computational overhead of each model (frame rates are computed on a single Nvidia\texttrademark~GTX 1080 GPU).}
%\scriptsize
%\vspace{-1mm}
%\begin{tabular}{l||r|r|r}
%  \hline
%  \textbf{Model} & \# of parameters &  Memory (MB) & Speed (FPS)  \\ \hline \hline
%  FCN8$_{CNN}$ & $69.744$ M & $837.1$ & $17.11$\\
%  FCN8$_{VGG}$ & $134.286$ M &$1630.0$ &$8.79$ \\
%  SegNet$_{CNN}$ & $2.845$ M & $46.3$ & $17.52$ \\
%  SegNet$_{ResNet}$ & $15.012$ M & $180.6$ & $10.86$ \\
%  UNet$_{GRAY}$ &$31.032$ M &$371.6$ &$20.13$ \\
%  UNet$_{RGB}$ & $31.033$ M & $372.6$ & $19.98$ \\
%  PSPNet$_{MobileNet}$ &  M &  &  \\
%  DeepLab$_{V3}$ &$41.254$ M  & $495.3$  &$16.00$ \\
%  \textbf{SUIM-Net} & $\mathbf{3.864}$ \textbf{M} & $\mathbf{46.8}$ & {$\mathbf{28.65}$} \\
%  \hline
%\end{tabular}
%\label{tab:usability}
%\end{table}%

SUIM-Net$_{RSB}$ and SUIM-Net$_{VGG}$ provides consistent and competitive performance for both region similarity and object localization. As Table~\ref{tab:quan_se} suggests, their $\mathcal{F}$ and \textit{mIOU} scores are within $5$\% margins of the respective top scores.        
The accuracy of semantic labeling and object localization can be further visualized in Figure~\ref{fig:qual}, which shows that the SUIM-Net-generated segmentation masks are qualitatively comparable to the ground truth labels. Although UNet$_{RGB}$, FCN8$_{VGG}$, and DeepLab$_{V3}$ achieve much fine-grained object contours, the loss is not perceptually significant. 
Moreover, as shown in Table~\ref{tab:usability}, SUIM-Net$_{RSB}$ operates at a rate of $28.65$ frames-per-second (FPS) on a Nvidia\texttrademark~GTX 1080 GPU, which is much faster than other SOTA models in comparison. Also, it is over $10$ times more memory efficient than UNet$_{RGB}$, FCN8$_{VGG}$, and DeepLab$_{V3}$. These computational aspects are ideal for its use in near real-time applications. 
However, its computational efficiency comes at a cost of lower performance margins and slightly poor generalization performance. 
To this end, with more learning capacity, SUIM-Net$_{VGG}$ achieves better object localization performance in general; it also offers an inference rate of $22.46$ FPS, which is still considerably faster than FCN8$_{VGG}$.

\begin{figure*}[t]
    \centering
    \includegraphics[width=0.99\linewidth]{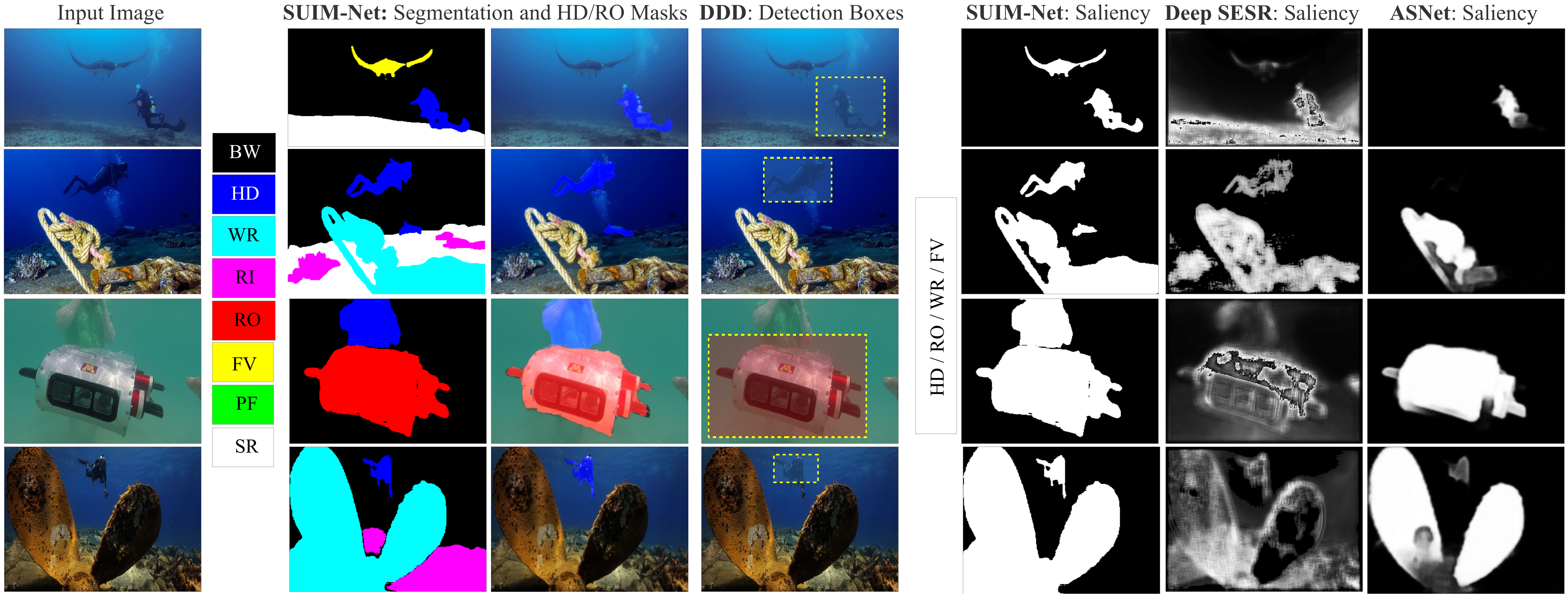}
    %\vspace{-1mm}
    \caption{SUIM-Net$_{VGG}$-generated segmentation masks are shown alongside the bounding box outputs of DDD~\cite{islam2018towards} for the detection of human divers and robots; also, their corresponding saliency masks are compared with the class-agnostic predictions of Deep SESR~\cite{islam2020sesr} and ASNet~\cite{wang2018salient}.}
    \label{fig:use}
\end{figure*}%

Figure~\ref{fig:use} further demonstrates the effectiveness of SUIM-Net$_{VGG}$-generated segmentation masks for fine-grained object localization in the image space. In particular, it compares the pixel-level detection of human divers and robots with the standard object detectors such as DDD~\cite{islam2018towards}. 
In addition to providing more precise object localization, SUIM-Net$_{VGG}$ incurs considerably fewer cases of missed detection, particularly in occluded or low-contrast regions in the image. Moreover, the additional semantic information facilitates much-improved saliency prediction, whereas the class-agnostic models such as Deep SESR~\cite{islam2020sesr} and ASNet~\cite{wang2018salient} concentrate on the high-contrast foreground regions only. 
These, among many others, are important use cases of the proposed SUIM dataset.
Further research efforts will be useful to explore the design and feasibility of various deep visual models for other application-specific attention modeling tasks. 

%% file: src/con.tex
\section{Conclusion}
Semantic segmentation of underwater scenes and pixel-level detection of salient objects are critically important features for visually-guided AUVs. The existing solutions are either too application-specific or outdated, despite the rapid advancements of relevant literature in the terrestrial domain. In this paper, we attempt to address these limitations by presenting the first large-scale annotated dataset for general-purpose semantic segmentation of underwater scenes. The proposed SUIM dataset contains 1525 images with pixel annotations for eight object categories: fish, reefs, plants, wrecks/ruins, humans, robots, sea-floor/sand, and waterbody background. We also provide a benchmark evaluation of the SOTA semantic segmentation approaches on its test set. 
Moreover, we present SUIM-Net, a fully-convolutional encoder-decoder model that offers a considerably faster run-time than the SOTA approaches while achieving competitive semantic segmentation performance. 
The delicate balance of robust performance and computational efficiency make SUIM-Net suitable for near real-time use by visually-guided underwater robots in attention modeling and servoing tasks. 
In near future, we plan to further utilize the SUIM dataset and explore various learning-based models for visual question answering and guided search; the subsequent pursuit will be to analyze their feasibility in underwater human-robot cooperative applications. 
%\vspace{5mm}

\section*{Appendix I: Data Augmentation Parameters}
{\small
We used the standard Keras libraries (\url{https://keras.io/preprocessing/image/}) for data augmentation in our work. The specific parameters are as follows: rotation range of $0.2$; width shift, height shift, shear, and zoom range of $0.05$; horizontal flip is enabled; and the rest of the parameters are left as default.
}

\section*{Appendix II: Relevant Source code}
{\small 
\begin{itemize}
    \item FCN variants: \url{https://github.com/divamgupta/image-segmentation-keras}
    \item BilinearUpsampling for FCN: \url{https://github.com/aurora95/Keras-FCN}
    \item SegNet variants and PSPNet: \url{https://github.com/divamgupta/image-segmentation-keras}
    \item UNet: \url{https://github.com/zhixuhao/unet}
    \item DeepLabv3: \url{https://github.com/MLearing/Keras-Deeplab-v3-plus/}
    \item Deep SESR: \url{https://github.com/xahidbuffon/Deep-SESR}
    \item ASNet: \url{https://github.com/wenguanwang/ASNet}
\end{itemize}
}

%\section*{ACKNOWLEDGMENT}
\section*{Acknowledgement}
This work was supported by the National Science Foundation grant IIS-\#1845364, the Doctoral Dissertation Fellowship (DDF) at the University of Minnesota and the Minnesota Robotics Institute (MnRI). We are grateful to the Bellairs Research Institute of Barbados for the field trial venue, and to the Mobile Robotics Lab of McGill University for data and resources.